# Learning Gentle Grasping from Human-Free Force Control Demonstration

Mingxuan Li[1], *Graduate Student Member, IEEE*, Lunwei Zhang[1], *Graduate Student Member, IEEE*, Tiemin Li[1], and Yao Jiang[1], *Member, IEEE*

*Abstract*—Humans can steadily and gently grasp unfamiliar objects based on tactile perception. Robots still face challenges in achieving similar performance due to the difficulty of learning accurate grasp-force predictions and force control strategies that can be generalized from limited data. In this article, we propose an approach for learning grasping from ideal force control demonstrations, to achieve similar performance of human hands with limited data size. Our approach utilizes objects with known contact characteristics to automatically generate reference force curves without human demonstrations. In addition, we design the dual convolutional neural networks (Dual-CNN) architecture which incorporates a physics-based mechanics module for learning target grasping force predictions from demonstrations. The described method can be effectively applied in vision-based tactile sensors and enables gentle and stable grasping of objects from the ground. The described prediction model and grasping strategy were validated in offline evaluations and online experiments, and the accuracy and generalizability were demonstrated.

## I. INTRODUCTION

Humans are skilled in grasping objects. Through tactile perception, humans can grasp objects of unknown shapes and textures stably and safely [1]-[3]. This behavior can be referred to as *gentle grasping*. As shown in Fig. 1, human grasp-lift actions always involve a similar process of four phases [1]. The regulation of force primarily occurs during the load phase: the object's movement is small, which results in minor skin deformation (usually submillimeter range [2]); However, humans can still quickly adjust force in a short period of time through subtle stimulation of slip and force. Two characteristics are highlighted in gentle grasping: 1) slip is prevented by maintaining the grasping force above the minimum force (required to prevent macro slip at a given moment); 2) the grasping force does not exceed the minimum force by much (typically no more than 60%) [4]. As a result, gentle grasping ensures a moderate force to lift an object stably without destroying the object or limiting dexterity [5].

Implementing gentle grasping in robotic systems is challenging [6]. Recently, the rapid development of tactile sensing technology has provided a promising avenue for inferring grasp stability [7], [8]. Array-type tactile sensors, such as magnetic-based e-skin [9], stand out for their integrability and replaceability. However, the dimensionality and density of the measured information are limited. Recently, vision-based tactile sensors are valued for offering rich tactile information [10]. Techniques based on photometric stereo [11] or marker displacement [12] can obtain high-resolution and multimodal tactile features, thus providing valuable deep details for stability assessment. Such sensors offer two ways to achieve gentle grasping: slip- and force-based approaches.

*This work was supported by the National Natural Science Foundation of China under Grant 52375017. [1]Mingxuan Li, Lunwei Zhang, Tiemin Li, and Yao Jiang (corresponding author) are with the Institute of Manufacturing Engineering, Department of Mechanical Engineering, Tsinghua University, Beijing 100084, China (e-mail: mingxuan-li@foxmail.com; zlw21@mails.tsinghua.edu.cn; litm@mail.tsinghua.edu.cn; jiangyao@mail.tsinghua.edu.cn).

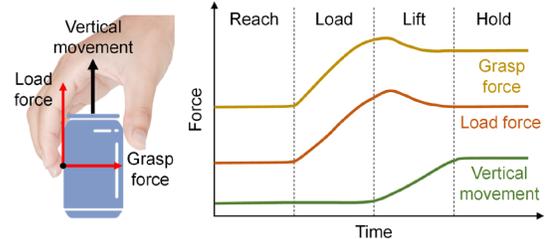

**Fig. 1.** The gentle grasping process of humans [1]. During the load phase, the object's vertical movement is very small, while the grasp force and load force have increased to near their peak values. The uncertainty of the object's properties causes overshoot in the regulation of forces, but humans can still stabilize it close to the minimum force within a short time of about 200 ms.

Slip-based methods ensure grasp stability by avoiding sliding [13]. Data-driven approaches, such as support vector machines (SVM) [14], the bimodal ConvLSTM network [15], normalized differential convolution (NDC) [16], and STNet [17], are widely employed to detect slip events. However, force control strategies based on slip classifiers typically adjust the grasp force after detecting sliding, which results in delayed responses to perturbations and short-term instability. One solution is to detect incipient slip, the transition from stable contact to macro slip, and use the stick ratio (the area ratio of the sticking region to the slipping region) to estimate the safety margin [18]-[20]. Yet, sui *et al.* demonstrated that using only the stick ratio to assess incipient slip is insufficient [21]. More detailed information regarding friction and force needs to be supplemented to enhance the quantification.

Force-based methods focus on direct indicators of slip or contact modeling [22]. Currently, force-related indicators are usually achieved through end-to-end methods, including the UNet model [23], the ShuffleNetV2 model [24], *etc*. However, since force distribution is often obtained through data-driven methods, the sensitivity of the training data always poses challenges for consistent data collection. Thus, physics-based modeling is required to decouple the tactile sensor's contact mechanics from the task-related tactile feedback model.

In addition to the requirement for contact mechanics, the combined demands for speed and stability also necessitate providing the robot with expert annotations for the entire process (i.e., a suitable grasp force control curve). The most common method for annotating data relies on manual human intuition [16]. However, designing the curve and collecting the data is often labor-intensive. Unlike detecting whether an object is slipping, expert data for gentle grasping tasks need to include the target force at each moment, which cannot be achieved manually. An automated method for generating reference force control curves is needed to achieve timely and effective grasp force control.

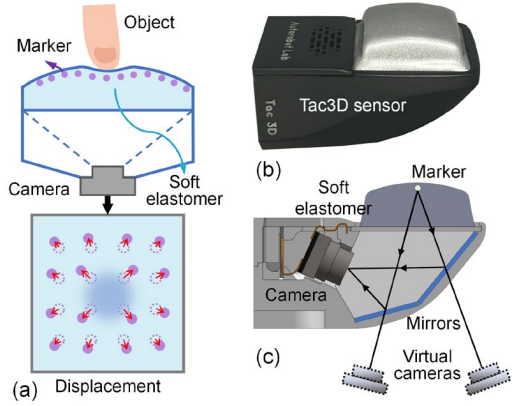

**Fig. 2.** (a) Marker displacement method used in vision-based tactile sensors. (b) Tac3D tactile sensor. (c) Virtual binocular vision system (VBVS).

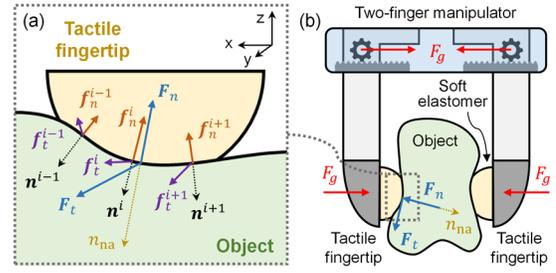

**Fig. 3.** (a) Contact model between a soft object and the tactile fingertip. (b) Grasping object with a two-finger manipulator.

This article proposes a strategy for learning gentle grasping through human-free force control demonstrations. This is a scheme for generating lightweight models. It utilizes pre-measured frictional properties of objects to construct an ideal force control demonstration, which is used for the robot to perform behavior cloning. This strategy can automatically generate expert trajectories without human involvement. Besides, to reduce training data while maintaining prediction accuracy, we incorporate a physics-based mechanical module into the training network. The force distribution obtained through the finite element method is not dependent on specific contact conditions, and is more sensitive to changes in grasping force and safety margins. As a result, gentle and stable grasping is achieved with a limited amount of data, and the model can generalize to unfamiliar objects.

The remainder of this article provides the following:
- The developed sensor and force reconstruction method.
- The control strategy for generating demonstrations and the network to learn from the demonstrations.
- Ablation studies and online experiments that evaluate the effectiveness of gentle grasping control.

## II. MATERIALS AND METHODS

### A. Tactile sensing and force reconstruction

We customized the vision-based tactile sensor, Tac3D [25], as the robot finger. Fig. 2(a) illustrates the principle of vision-based tactile sensors. When an external object contacts the soft elastomer of the sensor, the displacement of markers on the contact surface can be used to discretely sample contact deformations. The structure of Tac3D sensor is shown in Fig. 2(b). An internal camera (1920×1080 pixels) continuously captures the deformation of the elastomer at a frequency of 30 Hz. The surface of the elastomer is engraved with a 20×20 density marker pattern. Tac3D employs a virtual binocular vision system [26], as shown in Fig. 2(c). The key feature of this design is the use of mirrors to split the reflected light from the marker pattern into two paths. Similar designs can also be used to extend the sensor's effective range [27]. This solution ensures a balance between measurement accuracy, structural compactness, and synchronized triggering of the sensor.

Extracting distributed force information from deformation data has been shown to improve the assessment of grasping stability by suppressing features with minimal relation to contact and friction. Tac3D maps 3D deformation to 3D force distribution using finite element methods (FEM) [28], refined through various optimization calibration techniques [29]. Accurate force estimation with this approach relies on precise characterization of the gel's physical properties (e.g., Young's modulus and Poisson's ratio). Also, a new force mapping model needs to be constructed through FEM if the gel shape of the sensor changes. Despite these limitations, physics-based models are valued owing to being least affected by contact conditions and not requiring training for specific scenarios.

### B. Generation of Force Control Demonstrations

We consider the contact between a fingertip (i.e., the tactile sensor) and a deformable object, as shown in Fig. 3. For each position $p^i = (x^i, y^i, z^i)$ on the sensor's contact surface $S(x, y, z)$, the micro-element force denotes $\boldsymbol{f}^i = [f_x^i, f_y^i, f_z^i]^T$. Considering the contact deformation, $S$ is not flat. Therefore, the normalized normal direction of $S$ at $p^i$ can be calculate as

$$\boldsymbol{n}^i = [n_x^i, n_y^i, n_z^i]^T = \frac{1}{\|\nabla S\|} \cdot [\frac{\partial S}{\partial x}, \frac{\partial S}{\partial y}, \frac{\partial S}{\partial z}]^T. \quad (1)$$

Select the element patch at $p^i$, and the micro-element normal force $\boldsymbol{f}_n^i$ and tangential force $\boldsymbol{f}_t^i$ on it can be calculate as

$$\boldsymbol{f}_n^i = -(\boldsymbol{f}^i \cdot \boldsymbol{n}^i) \cdot \boldsymbol{n}^i, \quad (2)$$
$$\boldsymbol{f}_t^i = -\boldsymbol{f}^i + (\boldsymbol{f}^i \cdot \boldsymbol{n}^i) \cdot \boldsymbol{n}^i. \quad (3)$$

Therefore, according to Coulomb's friction, the condition for $p^i$ not to undergo local slip is:

$$\mu > \|\boldsymbol{f}_t^i\| / \|\boldsymbol{f}_n^i\|, \quad (4)$$

where $\mu$ is the coefficient of friction. Under the elastic contact assumption, this relationship also holds for soft objects.

We define the normalized average normal direction of $S$ as

$$\boldsymbol{n}_{\mathrm{na}} = \frac{\int_S \boldsymbol{n}^i \cdot dA}{\|\int_S \boldsymbol{n}^i \cdot dA\|}. \tag{5}$$

We consider $\boldsymbol{n}_{\mathrm{na}}$ as the normal of the equivalent contact plane. Therefore, the resultant normal force $\boldsymbol{F}_n$ and the resultant tangential force $\boldsymbol{F}_t$ of $S$ can be calculated as

$$\boldsymbol{F}_n = -\int_S (\boldsymbol{f}^i \cdot \boldsymbol{n}_{\mathrm{avg}}) \cdot \boldsymbol{n}_{\mathrm{avg}} \cdot dA, \tag{6}$$

$$\boldsymbol{F}_t = -\int_S [\boldsymbol{f}^i - (\boldsymbol{f}^i \cdot \boldsymbol{n}_{\mathrm{avg}}) \cdot \boldsymbol{n}_{\mathrm{avg}}] \cdot dA. \tag{7}$$

Since the contact surface is not flat, the contact stability cannot be directly described using the ratio of $F_t$ to $F_n$. The contact coefficient can be defined as an alternative [21], expressed as

$$cf = 1 - \frac{F_{MER,t}}{\mu \cdot F_{MER,n}}, \tag{8}$$

where $F_{MER,t}$ and $F_{MER,n}$ represent the micro-element resultant tangential and force normal force, respectively, as

$$F_{MER,n} = \int_S \|\boldsymbol{f}_n^i\| \cdot dA = \int_S \|(\boldsymbol{f}^i \cdot \boldsymbol{n}^i) \cdot \boldsymbol{n}^i\| \cdot dA, \tag{9}$$

$$F_{MER,t} = \int_S \|\boldsymbol{f}_t^i\| \cdot dA = \int_S \|\boldsymbol{f}^i - (\boldsymbol{f}^i \cdot \boldsymbol{n}^i) \cdot \boldsymbol{n}^i\| \cdot dA. \tag{10}$$

Note the distinction between $F_{MER,n}$ ($F_{MER,t}$) and $\boldsymbol{F}_n$ ($\boldsymbol{F}_t$): the latter refers to the component of the force vector integral, while the former represents the component of the force magnitude integral (not represents a physical force).

The contact factor ranges between 0 and 1. Ideally, all elements on the contact surface should be sticking. However, in practice, incipient slip always exists. Therefore, the closer $cf$ is to 0, the more regions are in or near the slip state, and the closer the contact state is to macro slip. Using $cf$ can effectively describe the contact stability and can be applied to grasping control. However, the friction coefficients of unfamiliar objects to be grasped are always unknown. Even if the friction coefficient is measured online through tactile sensing during the grasping, it usually takes tens of seconds [21]. Also, friction prediction in the incipient slip stage is often inaccurate (discussed in Section III-D).

Therefore, the contact factor is more suitable for generating demonstrations to guide the robot in learning to grasp. For objects with prior information, grasp force control curves can be generated by maintaining a certain contact factor to avoid macro slip. As a result, the robot can learn the force-following strategy to grasp objects with similar friction characteristics.

Based on the above discussion, we propose a strategy for generating force control demonstrations, as shown in Fig. 4(a). This strategy is not directly applied to grasping unknown objects but is used to collect force control data for grasping known objects, which the robot can then learn from. This concept is similar to imitation learning (also known as programming by demonstration) [30], but without human involvement, as the robot generates the expert data.

This article focuses solely on the two-finger parallel gripper as the end-effector. A basic strategy of gentle grasping is to control the grasp force $F_g$ during the load phase based on the currently measured load force and its increment, making these two forces match (i.e., ensuring stable contact). As shown in Fig. 3, $F_g$ can be expressed as

$$F_g = \int_S (\boldsymbol{f}^i \cdot \hat{\boldsymbol{z}}) \cdot dA = (\boldsymbol{F}_n + \boldsymbol{F}_t) \cdot \hat{\boldsymbol{z}}. \tag{11}$$

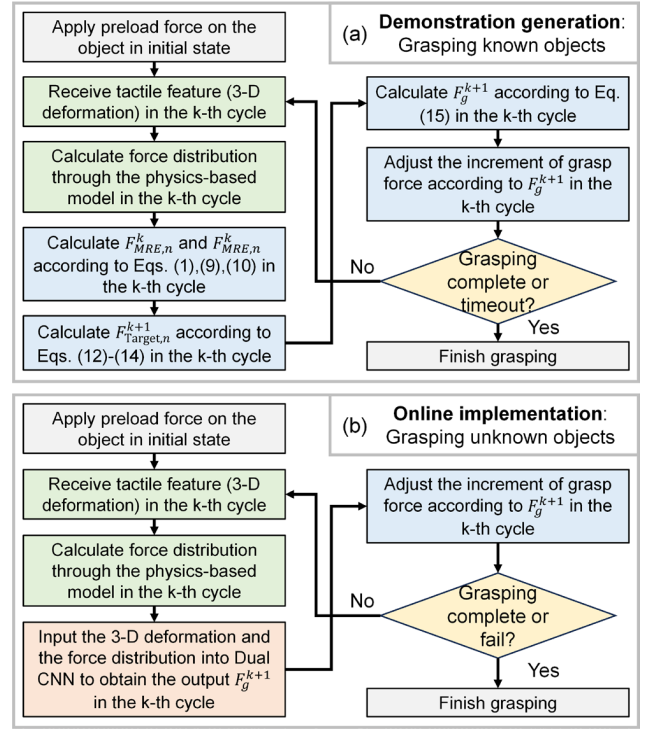

Fig. 4. (a) Flow chart of force-control demonstration generation. (b) Flow chart of online force control based on the Dual-CNN network.

During the transition between the load and lift phases, the object is already off the ground. At this point, the gripper should have reached the desired appropriate grasp force (i.e., the product of the safety margin and the minimum force).

The first step involves the meticulous measurement of the friction coefficient of the known objects in the training set. The method is as follows: the object is grasped with a constant force and lifted from the ground, then suspended in the air for 5 seconds. If no noticeable slip is observed during this process, the grasp force is slightly reduced. This process is repeated until the minimum grasping force is found. The ratio of the tangential force to the normal force at this point is precisely defined as the measured friction coefficient. Next, grasp force control is based on the following procedure: the robot contacts the object with a preload force $F_g^0 = 0.4N$. Then, during the load phase, the object is lifted slowly at a speed of 1 mm/s for 3 or 4 seconds. At frame $k$, the tactile sensor can measure the force distribution $\boldsymbol{f}$ and the contact geometry $\boldsymbol{S}$, so $F_{MRE,t}^k$ and $F_{MRE,n}^k$ can be calculated based on Eqs. (1) and (9)-(10). Then, at frame $k+1$, the robot needs to attempt to control the grasp force $F_g^{k+1}$ so that:

$$F_{\mathrm{Target},n}^{k+1} = \beta\mu^{-1} \cdot \max(F_{MRE,t}^k, F_m^{k-1}), \tag{12}$$

where $F_m^{k-1}$ represents the maximum value in $F_{MRE,t}^k$ and before frame $k+1$, and $F_{\text{Target},n}^{k+1}$ represents the target $F_{MER,n}^{k+1}$.

In Eq. 12, $\beta$ represents the safety margin. Initially, the grasping force is relatively small, and the tangential force increases rapidly, requiring a larger safety margin; as the grasping force gradually approaches the final target value and the tangential force increase rate decreases, a smaller safety margin is needed to prevent overshooting. Therefore, the safety margin is set as a time-dependent function:

$$\beta = \begin{cases} \beta_{\max}, & \text{if } t \geq t_m \\ \beta_{\min} + \dfrac{\beta_{\max} - \beta_{\min}}{1 + \exp(-k \cdot (t - t_{\text{bias}}))}, & \text{if } t < t_m \end{cases} \quad (13)$$

Eq. (13) uses a sigmoid function to non-linearly map the safety margin between $\beta_{\min}$ and $\beta_{\max}$. The parameters $\alpha$ and $t_{\text{bias}}$ control the shape of the sigmoid curve, thereby adjusting the strategy for selecting the safety margin. Here, $t$ and $t_m$ represent the camera frame count. The parameters were set as:

$$\beta_{\max} = 2, \ \beta_{\min} = 1.2, \ k = 0.1,$$
$$t_{\text{bias}} = 20, \ t_m = 180. \quad (14)$$

One remaining issue is the difference between $F_g^{k+1}$ and $F_{MRE,n}^k$, especially since the contact surface is not flat. Though the relationship between $F_g^{k+1}$ and $\boldsymbol{F}_n^{k+1}$ ($\boldsymbol{F}_t^{k+1}$) is determined by Eq. (11), the uncertain nature of soft object's deformation makes it impossible to predict the future relationship between $\boldsymbol{F}_n^{k+1}$ ($\boldsymbol{F}_t^{k+1}$) and $F_{MER,n}^{k+1}$ ($F_{MER,t}^{k+1}$) at frame $k$. An alternative solution is to dynamically estimate the linear relationship based on historical information (under the assumption of a unidirectional correlation [21]):

$$F_g^{k+1} = \begin{cases} F_g^k, & \text{if } F_{MER,n}^{k-1} = F_{MER,n}^k \\ F_g^k + (F_g^k - F_g^{k-1}) \cdot \dfrac{F_{\text{Target},n}^{k+1} - F_{MER,n}^k}{F_{MER,n}^k - F_{MER,n}^{k-1}}, \\ \quad \text{if } F_{MER,n}^{k-1} \neq F_{MER,n}^k \end{cases} \quad (15)$$

For the same object, force control curves implemented according to the above strategy may vary little. To further enhance the diversity of the training data, we randomly vary the initial preload force within the range of 0.3 to 0.5 and adjust the grasping positions on the same object. Additionally, to eliminate the influence of the lifting speed on force control, we uniformly sample from the original data sequence to adjust the time intervals between adjacent data points. Ultimately, the implemented demonstrations cover as many feasible paths as possible to ensure sufficient diversity in the training data.

### C. Target Force Prediction and Online Force Control

One advantage of prior-based force control demonstration is that it reduces reliance on temporal information. The generated reference force control follows a fixed strategy, so the collected data focuses on key contact states along the target path, avoiding trivial or unreachable states. In other words, subtle patterns and dependencies related to temporal sequences are already included in the demonstration. As a result, by sacrificing task generalization, the training process can focus on extracting spatial features. This approach is suited for small-sample, lightweight models, as it eliminates the need for continuous frame training, enabling effective performance with less data and simpler networks.

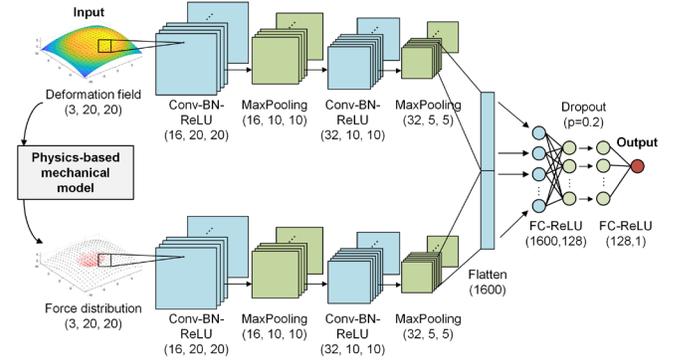

Fig. 5. Diagram of the Dual-CNN network for target force prediction. Conv-BN-ReLU: convolution layer, batch normalization layer, and ReLU activation layer. FC-ReLU: fully connected layer and ReLU activation layer.

The goal of training is to predict the target grasping force to be controlled at the next moment based on the data collected by the tactile sensor at the current moment. The overall architecture of the proposed target force prediction network is shown in Fig. 5. The input information to the network is the contact deformation measured by Tac3D (stored in the form of a 3-D coordinate field). We first map the displacement field to the 3-D distributed force field based on the method described in Section II-A. The region where the normal force exceeds the threshold value (set to 0.01 mm) is considered as the contact region. We suppress the values of the distributed force outside the contact region using graphics operations to reduce measurement noise at non-contact region.

At the $k$-th moment, the coordinate and force samples are coordinated in the same format and fed into a Dual-CNN to extract features. The main reason for using CNNs is to ensure translation invariance. The CNN module comprises a stack of 2 layers of Conv-BatchNorm-ReLU (CBR) and two pooling layers. The output features of the two CNN sub-networks are flattened and spliced, and the concatenated features are mapped to the force prediction using a multi-layer perceptron (MLP). A dropout operation with a rate of 0.2 is applied.

The online grasping force control is implemented based on the control method shown in Fig. 4(b). First, a preload force of 0.4 N is applied to the object. During the load phase, the robot slowly lifts the object at a speed of 1 mm/s. Meanwhile, the data collected by the Tac3D sensor is fed into the Dual-CNN model to generate the current target grasping force. The gripper adjusts and maintains the grasp force until the sensor gets new data. The control lag caused by algorithm processing time and communication delay does not exceed 50 ms. When the time reaches 4 seconds, or if the tangential force does not change significantly within 0.2 seconds, the final target grasp force is considered to have been reached. Finally, in the lift phase, the tactile sensor no longer collects data, and the object is lifted at a high speed of 10 mm/s (for grasping efficiency) while maintaining the final force from the load phase.

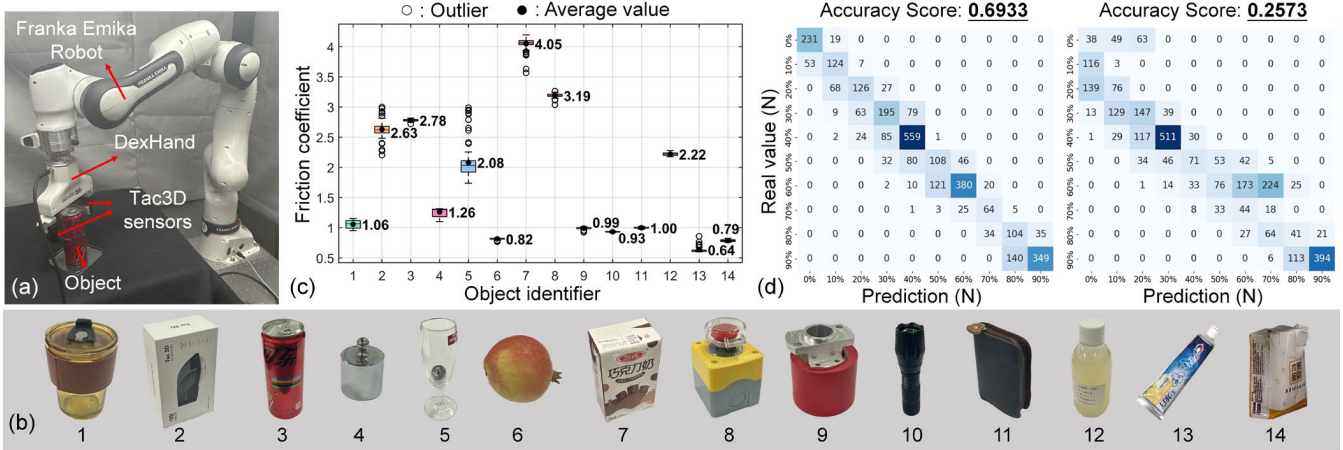

**Fig. 6.** (a) Experimental platform. (b) Objects used for data collection. Objects 1-7 were used to train and test the model, 8-12 were used for online evaluations, and 13-14 were used for comparative experiments. (c) Mean and standard deviation of the friction coefficient estimated for each object. The horizontal coordinate is the serial number of each object in Fig. 6(a). (d) Confusion matrixes. Left: Dual-CNN (trained with deformation and force). Right: CNN (trained with deformation only). The more the data is concentrated on the diagonal, the better the prediction (accuracy score) is.

## III. EXPERIMENTS AND RESULTS

### A. Experimental Setup

The experimental platform is shown in Fig. 6(a). We used a Franka Emika robot arm and a custom-designed two-finger gripper with good force-following performance. Considering computational efficiency, only the readings from one Tac3D sensor were used. In the offline evaluation, all models were implemented by PyTorch and trained on a laptop with a 2.30 GHz Intel i7-12700H processor and a GeForce 3060 GPU (6 GB). For the online experiments, the whole pipeline was run on an Ubuntu 18.04 PC with a 3.6 GHz AMD Ryzen 5 processor and a GeForce 1060 GPU (3 GB).

Fourteen household items with different shapes, materials, and stiffness were used to collect data, as shown in Fig. 6(b). We added metal pieces on top of or inside lighter objects (e.g., objects 5, 9, and 13) to ensure they could not be easily lifted. Fig. 6(c) shows the measured friction coefficients of these objects using the method described in Section II-B. One hundred grasps were conducted for each object, and each process involved 90-120 tactile frames. 81,371 sets of 3-D deformation fields (inputs) and target forces (labels) were collected. We used 80% of this data for training and 20% for testing. The training involves 50 epochs with a batch size of 16. The network is trained using stochastic gradient descent (SGD) with a learning rate of $10^{-6}$, and the Mean Squared Error (MSE) loss function and Adam optimizer are used.

### B. Offline Evaluation

The result of the ablation study is shown in Table 1. We compared the Dual-CNN structure with a model containing only a single active CNN pathway. The difference is that the latter removes the physics-based force reconstruction module (i.e., only the CNN path with the 3-D coordinate field is used). As a result, the mean squared error loss (MSE Loss) increased by nearly seven times. Also, we examined the impact of kernel size and depth on the Dual-CNN, where the added layer followed the Conv-BN-ReLU (64, 5, 5) pattern. The results indicate that the impact of parameter and structural changes on the Dual-CNN is limited. Thus, the improvement achieved primarily stems from the physics-based mechanics module.

**TABLE I**
PERFORMANCE COMPARISON OF DIFFERENT METHODS

| Models | MSE test loss / $N^2$ |
|---|---|
| CNN (trained with deformation only) | 0.0236 |
| Dual-CNN (3×3 kernels) | 0.0037 |
| Dual-CNN (5×5 kernels) | 0.0036 |
| Dual-CNN (decrease one layer) | 0.0039 |
| Dual-CNN (increase one layer) | 0.0054 |

We divided the true values of the target grasping force into ten intervals to evaluate the model's performance across the entire range, from 0% to 100% of the maximum prediction result. The confusion matrixes are shown in Fig. 6(d). By comparing the diagonal trends in the figure, it can be seen that the target force estimated by the proposed model is positively correlated with the ground truth, and the model with force reconstruction module performs better. The results above demonstrate that the force reconstruction module effectively improves the accuracy of target grasping force prediction.

### C. Online Evaluation

We conducted 30 online grasping experiments on objects 8 to 12, respectively [see Fig. 7(a)]. The objects were unknown to the robot, and no safety margins were artificially set. Therefore, the targe force was automatically determined by the proposed model. Fig. 7(b) shows all experiments' load-force/grasp-force states during the load phase. The discrete points of the same color represent the force states during the same grasping trial. We define safety boundaries and stability boundaries to describe the upper limit to prevent excessive grasp forces and a lower limit to prevent falling events. Both boundaries are determined based on the pre-measured friction coefficients but are not known as the prior information.

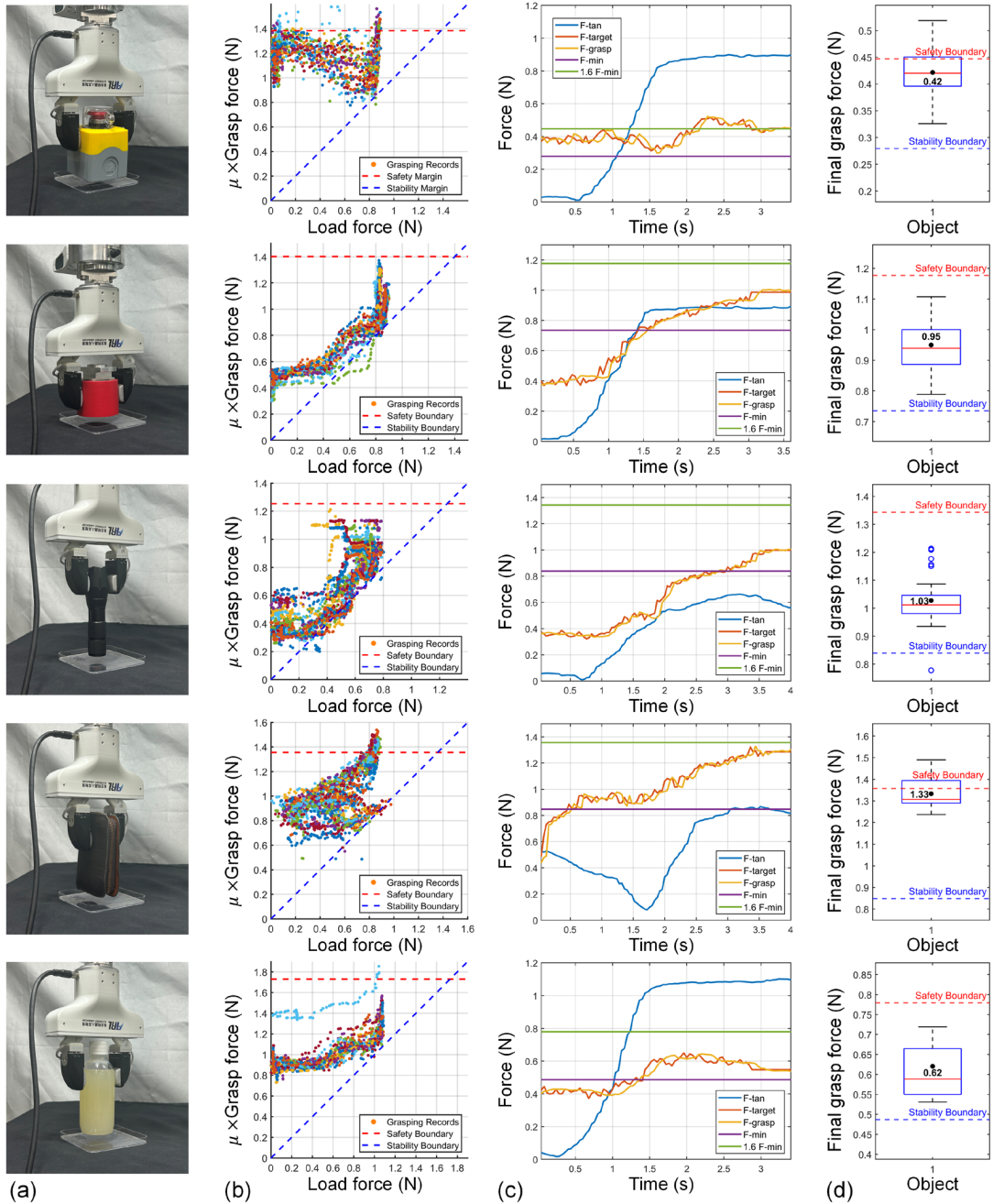

**Fig. 7.** Online evaluation of grasp force control. (a) Schematic of the grasping process. (b) Evaluation of load-force/grasp-force status during the whole process. The safety boundary is the product of the minimum grasp force required to pick up the object and a safety margin (set to 1.6 in the experiments), and the stability boundary is defined as the minimum force corresponding to the current load force. (c) Force curve during the whole process. F-tan: load force. F-target: target grasp force (network output). F-grasp: actual grasp force. F-min: minimum grasp force to lift the object. (d) Statistical results of the final grasp force.

Out of all test cases, only object 9 experienced a single drop case (corresponding to the green discrete points in Fig. 7(b) for that particular grasping attempt). In most cases, the force state can be maintained between the safety and stability boundaries during the process. The randomly selected force curves are shown in Fig. 7(c). The robot can quickly complete the load phase for different objects within 2~3 seconds. Despite some disturbance and delay, the proposed method still ensures that the change in grasp force remains synergistic with the shift in load force. Fig. 7(d) shows the final grasp force distribution for all sub-attempts. Most tests meet the requirements for gentle and stable grasping or slightly exceed the safety margin. Since actual objects cannot be broken by less than twice the minimum grasp force, such results are sufficient for the gentle grasping of most household objects.

Besides, Fig. 7(b) shows that object properties affect the performance. Regarding shape, the force control performance for cylindrical objects (objects 9 and 12) is better compared to block-shaped objects (objects 8 and 11), which stems from the imbalance in handling object shapes in the training set. For lighter objects (e.g., object 8), the ideal grasping force was close to the initial preload force, resulting in more pronounced overshooting in the midsection of the force regulation curve. In contrast, heavier objects (e.g., object 10) require higher force, increasing the risk of the force state approaching the safety boundary. Regarding texture, although differences in friction coefficients have a minor impact on the method, rough surfaces reduce the accuracy of force prediction (e.g., object 11). As for the grasp pose, since the training set only covers translational slip conditions, when significant pivot rotation occurs during grasping (e.g., object 9), the predicted force tends to be smaller, and the object is more likely to slip.

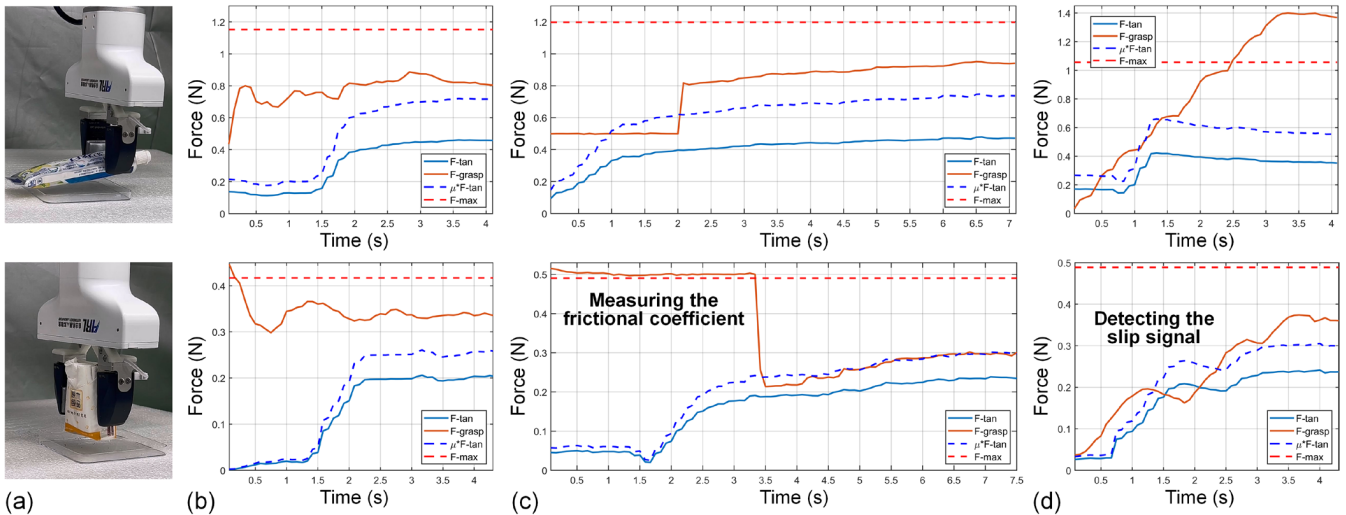

**Fig. 8.** Comparative experiment of different methods. (a) Schematic of the grasping process. (b) Grasping based on the proposed method. (c) Grasping based on online friction coefficient measurement [21]. (d) Grasping based on a simple slip detection model.

*D. Comparative Experiment*

The proposed method is compared with two typical force regulation measurement methods [see Fig. 8]. The first approach plans the force by estimating the friction coefficient online during the incipient slip, according to the details in [21] (with a preload force of 0.5 N). For the second one, we employ a slip feedback control strategy similar to [15] in a slip detector based on [19], [20]. Fig. 8(b)-(d) show the force curves for grasping objects with these methods, respectively. The blue dashed line represents the lower limit of the grasp force at each moment. The red dashed line represents the product of the minimum grasp force and a safety margin of 1.6, which indicates the acceptable maximum grasping force.

The time required for force control in the second method is significantly longer (approximately 0.75 times) than that of the proposed method [see Fig. 8(c)]. During the incipient slip, the tactile features primarily come from sub-millimeter-scale skin deformations [2], which are less reliable due to measurement errors. Thus, the tactile sensor needs a certain preload process to collect sufficient data to fit the frictional coefficient. According to [21], a complete process typically takes over 20 seconds. Also, the results indicate that bias in the estimation of the frictional coefficient can lead to forces approaching the lower boundary during the grasping process.

Fig. 8(d) illustrates the slip-based force control results. The grasp force curve exhibits a stepwise increase, as the controller raises the grasping force whenever slip is detected to prevent further slip propagation. The issue, however, lies in the fact that when a slip occurs, the object's state is inherently unstable, causing the force curve to fall below the lower boundary at certain points. In other words, a simple slip detection model cannot ensure that the contact surface remains during the incipient slip phase. Besides, without prior information of the object, inappropriate force increments may cause the grasp force to exceed the safety boundary. The relationship between slip detection and force control requires either prior knowledge of the object or data-driven learning approaches.

Compared to the other two methods, the advantage of force planning based on demonstrations lies in its adaptive adjustment. Even if the predicted force in the previous frame is too large or too small, the timely feedback provided by tactile sensing is reflected in the input, resulting in a force closer to the optimal value. Also, the heuristic method can offer prior knowledge about the task. As a result, the described method eliminates the need to measure the object's characteristics online and can shorten the loading phase duration.

## IV. CONCLUSION

This article proposes a method based on learning force reference trajectories for achieving stable and gentle grasping. We focus on how to generate ideal reference cases for grasp force control, and how to use a lightweight network in conjunction with physics-based force reconstruction. The strategy can predict the target force and generalize to unknown objects. Ablation experiments and online tests show the application of the described approach.

The limitation of the described method is the dependence on the specific grasping action. The idea of imitation learning, while helping to avoid the complexity of learning on long time-series information, makes the process applicable only to lifting an object vertically from the ground. For objects off the center of gravity that may rotate during the grasping, the network may need to be retrained to include more data related to rotational slip effects. For objects with particularly low coefficients of friction or soft structure, the method may not be able to adjust to the final target gripping force in a very short period of time. This phenomenon has also been mentioned in other works [24]. For example, when trying to grasp a frozen orange (friction coefficient of about 0.3) or a soft plush toy (the elastic modulus is lower than that of the sensor's elastomer), this method fails due to out-of-distribution.

Future work will consider the introduction of network modules capable of processing temporal information and explore techniques for generating differentiated force control demonstrations. Although this article focuses on applying lightweight models with limited data, learning from demonstrations without human input can also benefit from larger-scale frameworks such as vision transformers (ViT) for more complex tasks. Additionally, integrating visual and tactile information [31] or using reinforcement learning trained in simulation and real-world deployment [32] can process spatial-temporal features and improve accuracy by expanding the size and diversity of the training dataset.

REFERENCES

[1] R. S. Johansson, and J. R. Flanagan, "Coding and use of tactile signals from the fingertips in object manipulation tasks," *Nature Reviews Neuroscience*, vol. 10, no. 5, pp. 345–359, 2009.
[2] N. Afzal *et al.*, "Submillimeter lateral displacement enables friction sensing and awareness of surface slipperiness," *IEEE Trans. Haptics*, vol. 15, no. 1, pp. 20–25, Jan./Mar. 2022.
[3] B. P. Delhaye, F. Schiltz, F. Crevecoeur, J.-L. Thonnard, and P. Lefèvre, "Fast grip force adaptation to friction relies on localized fingerpad strains," *Science Advances*, vol. 10, no. 3, 2024, Art. no. eadh9344.
[4] A. M. Hadjiosif and M. A. Smith, "Flexible control of safety margins for action based on environmental variability", *Journal of Neuroscience*, vol. 35, no. 24, pp. 9106-9121, 2015.
[5] T. Bi, C. Sferrazza and R. D'Andrea, "Zero-shot sim-to-real transfer of tactile control policies for aggressive swing-up manipulation", *IEEE Robotics and Automation Letters*, vol. 6, no. 3, pp. 5761-5768, 2021.
[6] Z. Xie, X. Liang, and C. Roberto, "Learning-based robotic grasping: A review," *Frontiers in Robotics and AI*, vol. 10, 2023, Art. no. 1038658.
[7] Nathan F. Lepora, "The future lies in a pair of tactile hands" *Science Robotics*, vol. 9, no. 91, 2024, Art. no. eadq1501.
[8] T. Li *et al.*, "A comprehensive review of robot intelligent grasping based on tactile perception," *Robotics and Computer-Integrated Manufacturing*, vol. 90, 2024, Art. no. 102792.
[9] R Bhirangi *et al.*, "Anyskin: Plug-and-play skin sensing for robotic touch," Sep. 2024, arXiv:2409.08276.
[10] S. Zhang et al., "Hardware technology of vision-based tactile sensor: A review," *IEEE Sensors Journal*, vol. 22, no. 22, pp. 21410–21427, Nov. 2022.
[11] M. K. Johnson and E. H. Adelson, "Retrographic sensing for the measurement of surface texture and shape", in *2009 IEEE Conference on Computer Vision and Pattern Recognition (CVPR)*, 2009, pp. 1070-1077.
[12] M. Li, L. Zhang, T. Li, and Y. Jiang, "Marker displacement method used in vision-based tactile sensors—from 2D to 3D: A review," *IEEE Sensors Journal*, vol. 23, no. 8, pp. 8042-8059, 2023.
[13] S. Dong, D. Ma, E. Donlon, and A. Rodriguez, "Maintaining grasps within slipping bounds by monitoring incipient slip," in *2019 IEEE International Conference on Robotics and Automation (ICRA)*, May 2019, pp. 3818–3824.
[14] J. W. James and N. F. Lepora, "Slip detection for grasp stabilization with a multifingered tactile robot hand," *IEEE Transactions on Robotics*, vol. 37, no. 2, pp. 506–519, Apr. 2021.
[15] S. Cui, S. Wang, R. Wang, S. Zhang, and C. Zhang, "Learning-based slip detection for dexterous manipulation using GelStereo sensing." *IEEE Transactions on Neural Networks and Learning Systems*, pp. 1-10, 2023.
[16] Z. Zhao, W. He, and Z. Lu, "Tactile-based grasping stability prediction based on human grasp demonstration for robot manipulation," *IEEE Robotics and Automation Letters*, vol. 9, no. 3, pp. 2646-2653, March 2024.
[17] J. Lu, B. Niu, H. Ma, J. Zhu and J. Ji, "STNet: Spatio-temporal fusion-based self-attention for slip detection in visuo-tactile sensors," in *2024 IEEE International Conference on Robotics and Automation (ICRA)*, 2024, pp. 3051-3056.
[18] W. Chen, H. Khamis, I. Birznieks, N. F. Lepora, and S. J. Redmond, "Tactile sensors for friction estimation and incipient slip detection—Toward dexterous robotic manipulation: A review," *IEEE Sensors Journal*, vol. 18, no. 22, pp. 9049–9064, Nov. 2018.
[19] R. Sui, L. Zhang, T. Li, and Y. Jiang, "Incipient slip detection method with vision-based tactile sensor based on distribution force and deformation," *IEEE Sensors Journal*, vol. 21, no. 22, pp. 25973–25985, Nov. 2021.
[20] M. Li, Y. H. Zhou, T. Li, and Y. Jiang, "Incipient slip-based rotation measurement via visuotactile sensing during in-hand object pivoting," in *2024 IEEE International Conference on Robotics and Automation (ICRA)*, 2024, pp. 17132-17138.
[21] R. Sui, L. Zhang, Q. Huang, T. Li, and Y. Jiang, "A novel incipient slip degree evaluation method and its application in adaptive control of grasping force," *IEEE Transactions on Automation Science and Engineering*, pp. 1-10, 2023.
[22] M. Costanzo, G. De Maria, and C. Natale, "Slipping control algorithms for object manipulation with sensorized parallel grippers," in *2018 IEEE International Conference on Robotics and Automation (ICRA)*, 2018, pp. 3675–3681.
[23] P. Griffa, C. Sferrazza, and R. D'Andrea, "Leveraging distributed contact force measurements for slip detection: A physics-based approach enabled by a data-driven tactile sensor," in *2022 IEEE International Conference on Robotics and Automation (ICRA)*, May 2022, pp. 4826–4832.
[24] D. -J. Boonstra, L. Willemet, J. Luijkx and M. Wiertlewski, "Learning to estimate incipient slip with tactile sensing to gently grasp objects," in *2024 IEEE International Conference on Robotics and Automation (ICRA)*, 2024, pp. 16118-16124.
[25] L. Zhang, Y. Wang, and Y. Jiang, "Tac3D: A novel vision-based tactile sensor for measuring forces distribution and estimating friction coefficient distribution," Feb. 2022, arXiv:2202.06211.
[26] Y. Wang, L. Zhang, T. Li, and Y. Jiang, "A model-based analysis-design approach for virtual binocular vision system with application to vision-based tactile sensors," *IEEE Transactions on Instrumentation and Measurement*, vol. 72, p. 5010916, 2023.
[27] Y. Ma, J. A. Zhao and E. Adelson, "GelLink: A Compact Multi-phalanx Finger with Vision-based Tactile Sensing and Proprioception," in *2024 IEEE International Conference on Robotics and Automation (ICRA)*, 2024, pp. 1107-1113.
[28] L. Zhang, T. Li, and Y. Jiang, "Improving the force reconstruction performance of vision-based tactile sensors by optimizing the elastic body," *IEEE Robotics and Automation Letters*, vol. 8, no. 2, pp. 1109–1116, Feb. 2023.
[29] M. Li, L. Zhang, Y. H. Zhou, T. Li and Y. Jiang, "EasyCalib: Simple and low-cost in-situ calibration for force reconstruction with vision-based tactile sensors," *IEEE Robotics and Automation Letters*, vol. 9, no. 9, pp. 7803-7810, 2024.
[30] K. Yamane, Y. Saigusa, S. Sakaino and T. Tsuji, "Soft and rigid object grasping with cross-structure hand using bilateral control-based imitation learning," *IEEE Robotics and Automation Letters*, vol. 9, no. 2, pp. 1198-1205, Feb. 2024.
[31] H.Qi, A.Kumar, R.Calandra, Y. Ma, and J.Malik, "In-hand object rotation via rapid motor adaptation," in *Proc. Conf. Robot Learn.*, 2022, pp. 1722-1732.
[32] Y.Han *et al.*, "Learning generalizable vision-tactile robotic grasping strategy for deformable objects via transformer," *IEEE/ASME Trans. Mechatron.*, Jun. 07, 2024, doi: 10.1109/TMECH.2024.3400789.